# A Novel and Accurate BiLSTM Configuration Controller for Modular Soft Robots with Module Number Adaptability

Zixi Chen[1]*, Matteo Bernabei[1], Vanessa Mainardi[1], Xuyang Ren[2], Gastone Ciuti[1], and Cesare Stefanini[1]

## Abstract

Modular soft robots have shown higher potential in sophisticated tasks than single-module robots. However, the modular structure incurs the complexity of accurate control and necessitates a control strategy specifically for modular robots. In this paper, we introduce a data collection strategy and a novel and accurate bidirectional LSTM configuration controller for modular soft robots with module number adaptability. Such a controller can control module configurations in robots with different module numbers. Simulation cable-driven robots and real pneumatic robots have been included in experiments to validate the proposed approaches, and we have proven that our controller can be leveraged even with the increase or decrease of module number. This is the first paper that gets inspiration from the physical structure of modular robots and utilizes bidirectional LSTM for module number adaptability. Future work may include a planning method that bridges the task and configuration spaces and the integration of an online controller.

## Keywords

Data-driven control, bidirectional LSTM, configuration control, modular soft robot

[1] Biorobotics Institute and Department of Excellence in Robotics and AI, Scuola Superiore Sant'Anna, Pisa, Italy.
[2] Multi-scale Medical Robotics Centre and Chow Yuk Ho Technology Centre for Innovative Medicine, The Chinese University of Hong Kong, Hong Kong, China
* Corresponding author. Email: Zixi.Chen@santannapisa.it

# 1. Introduction

Soft robots have been widely applied in numerous areas due to their softness and high degrees of freedom (DOFs,) and a variety of soft robots have been designed for different applications. For example, concentric tube robots[1] are leveraged in robot-assisted surgery, especially minimally invasive surgery (MIS.) Also, cable-driven robots have been validated in cardiothoracic endoscopic surgery[2]. Due to their safety, researchers include pneumatic robots in recovery devices[3] and assistive robots[4]. A soft six-legged untethered robot can walk with the actuation of fluid-driven actuators[5]. Soft robots can also be employed as biorobots to mimic the behaviors of animals like fish[6], octopus[7–9], and elephant[10]. Overall, the robot community has taken advantage of many categories of soft robots in various research topics[11–14].

Modular soft robots have shown unique capabilities compared with other soft robots. Multiple modules endow the robot system with reconfiguration[15] and multiple choices of module numbers. In this case, they are flexible and can meet the requirements of different tasks. Compared to single-module robots, modular robots have more degrees of actuation (DOAs) and, therefore, more active DOFs. Modular robots can provide larger working spaces from the views of both kinematics and dynamics. Thanks to these properties, modular robots can achieve complex manipulations[16,17].

Moreover, shape control can be implemented on modular soft robots. Most control targets of single modular robots are only end positions[18–20]. Once the end positions are controlled, all the robot states, including the end orientations and robot shapes, are decided and depend on the end positions. But in many cases, the end positions and orientations of modular soft robots are independent. The robot can keep the end positions/orientations invariant and change the end orientations/positions[21]. By controlling the robot shapes, modular robots can avoid obstacles[22] and cross holes[23].

However, along with these advantages, modular soft robots encounter difficulties in accurate control. In addition to the nonlinearity and hysteresis of soft robots, such modular systems have more input variables than single-module robots, complicating the dynamics model. As shown in Figure 1-(A), the motion of each module will directly affect the configurations of two adjacent modules and indirectly affect the whole robot system. Moreover, the configurations of these modules decide the robot states with different levels. Even a tiny vibration on the modules near the base will crucially change the robot shape due to the accumulated motion along the modular robot, but the modules near the end show lower importance. In summary, although this kind of robot has high potential, the complex model necessitates a control strategy specifically for modular soft robots.

Some researchers have proposed approaches to utilize modular soft robots in manipulation. A series of research has been carried out on a modular pneumatic robot[16,21,24–27]. Due to the high model complexity, data-driven approaches are leveraged to achieve path tracking[24], end pose control[21], and sophisticated manipulation tasks like opening a drawer, turning a handwheel, and drawing a line on a paper[27]. Of note, the modular robot in [21] keeps the end position/orientation invariant while changing the end orientation/position, which can be achieved exclusively by modular robots. A modular robot named Robostrich arm[28] is applied to mimic the behavior of an ostrich and achieve the reaching task in a narrow space. However, the applied reinforcement learning methods are time-consuming.

Also, some physical models for modular soft robots are proposed. Piecewise constant curvature (PCC) and the Cosserat approach are two of the most widely applied physical models in soft robots[29]. Neural networks are employed in [30] for measurement, and PCC is employed for modular soft robot modeling. Kinematics models of modular soft robots are investigated in [31,32], also based on PCC. The discrete Cosserat approach is utilized to simulate multisection soft manipulator dynamics[33]. However, with the improvement of module number, the complexity of the physical model increases dramatically, and the most sophisticated robot in these physical model works has only three modules[31], whose model is only validated in simulation.

This paper aims to introduce a bidirectional long short-term memory (biLSTM) configuration controller for modular soft robots with module number adaptability. Instead of proposing one controller for each module number, our biLSTM network can work as a module from the view of the controller and is available to robots with different module numbers. With such a controller, modular soft robots can fulfill some tasks that single-module robots cannot achieve with relatively low errors.

The contributions of this paper are summarized as follows:
- We propose a novel and accurate biLSTM controller for modular soft robots that aims to control the module configurations and can be applied to robots with different module numbers.
- We introduce a data collection method specifically for modular robots to reach states away from the resting state.
- Experiments are carried out to validate the proposed controller on simulated cable-driven robots and real pneumatic robots. Configuration control tasks on robots with two, three, four, and six modules are achieved with the help of this controller.

## 2. Methods

### 2.1. BiLSTM Controller

To propose a controller for modular robots, we first analyze their structures. As shown in Figure 1-(B), these modules share a similar shape, stiffness, and other physical properties. Their states can be represented by the same kind of configurations. They are connected in sequence, and each module will interact with two adjacent modules. In this module sequence, modules have different levels of impact on the robot motion.

Based on the above analysis, we leverage biLSTM as the module configuration controller. Due to the hysteresis in soft robot space sequences, recurrent neural networks (RNNs) have been applied for control because such networks can address issues related to sequences. Considering the bidirectional influence between modules, we apply a bidirectional RNN, biLSTM, shown in Figure 1-(C) for configuration control. To train the biLSTM units and control modular robots, we can connect biLSTM units that share the same module number with robots, even if the training and controlled robots have different module numbers. As far as we know, this is the first paper that considers the space sequence of soft robots and utilizes RNN to solve this problem for module number adaptability.

The diagram of one biLSTM unit is shown in Figure 1-(D). In module II, the forward LSTM network takes the cell state $C_{fI}$ and the hidden state $h_{fI}$ from module I as input, and provides the cell state $C_{fII}$ and the hidden state $h_{fII}$ of this unit to module III. Meanwhile, the backward LSTM network takes the cell state $C_{bIII}$ and the hidden state $h_{bIII}$ from module III as input, and provides the cell state $C_{bII}$ and the hidden state $h_{bII}$ of this unit to module I. In addition to these states, the module label $n_{II}$, the desired module state $S_d$, and previous actions and module states $S_{II0}, A_{II0}, S_{II1}, \cdots$ are fed into these two networks. Their outputs are concentrated and fed into a fully connected layer for actuation estimation.

The module labels are used to infer the module position in the sequence, and the label of module $i$ can be denoted as

$$n_i = \frac{2(i-1)}{n_{sum} - 1} - 1, \tag{1}$$

where $n_{sum}$ is the amount of modules in this robot. The range of the labels is [-1,1], and a large label represents that this module is away from the robot base.

Each LSTM unit can be seen as

$$\begin{aligned} f &= sig(W_f \cdot [h_{-1}, x] + b_f), \\ i &= sig(W_i \cdot [h_{-1}, x] + b_i), \\ C &= f \times C_{-1} + i \times tanh(W_c \cdot [h_{-1}, x] + b_c), \\ o &= sig(W_o \cdot [h_{-1}, x] + b_o), \\ h &= o \times tanh(C), \end{aligned} \tag{2}$$

where $f, i, C, o,$ and $h$ denote the forget gate, input gate, cell state, output gate, and hidden state of this LSTM unit, respectively. $h_{-1}$ and $C_{-1}$ are the hidden state and cell state provided by the other unit. $x$ denotes the LSTM input, and $W_*$ and $b_*$ denote the weight and bias parameters of the corresponding states. $\times$ is the Hadamard product operator, and $sig$ is the sigmoid function.

## 2.2. Data Collection

To train a neural network as a robot controller, it is necessary to collect a dataset in simulation or the real world. Generally, the robot will be actuated randomly to generate this dataset, and such a dataset can cover the whole working space in most cases, as shown in Figure 2-(A). However, due to the multiple modules, the modular robot tends to vibrate near the resting state instead of exploring and reaching the edge of the working space. Therefore, we propose a data collection approach specifically for modular robots, as shown in Figure 2-(B), which takes a soft robot with three modules as an example.

In the first one-third of the data collection process, all the modules are actuated with the same random action sequence $a_{a1}$, and the robot can move far away from the resting state. In the following one-third process, the end module is actuated with a random action sequence $a_{b3}$, and the other two modules are actuated with a different random action sequence $a_{b1}$. In the final one-third process, three modules are actuated with different random action sequences $a_{c1}$, $a_{c2}$ and $a_{c3}$, respectively. This data collection approach allows the modular robot to reach a larger space than the traditional method.

## 3. Results

## 3.1. Experimental Setup

To validate our biLSTM controller and data collection strategy, we include simulation cable-driven robots and real pneumatic robots in our experiments.

The simulation robot is based on PyElastica[34,35], as shown in Figure 3-(A). The real robot system in Figure 3-(B) is composed of pneumatic modules covered by origami structures. Each module has two opposite chambers connected to valves (Camozzi K8P-0-E522-0) under the control of the Arduino MEGA control board. Each module is connected to the other modules with 3D-printed connectors. Optical track makers are fixed on the connectors, and NDI Polaris tracks the module motion. The real experimental setup runs at 2.5 Hz. Detailed information about the setup, soft module structure, and connectors is included in Figure S1.

To represent the module configuration, we utilize the unit vector of the module end relative to the module base, as shown in Figure 3-(C), (D). In this case, the simulation and real module configurations are denoted by $[v_x, v_y, v_z] \in R^3$ and $[v_x, v_y] \in R^2$, and the range of each value is [-1,1]. Each module has four cables in simulation, and we utilize two actuation values, $a_0, a_1 \in [-1,1]$, to control them.

$$a_I = \max\{0, a_0\},$$

$$a_{II} = \max\{0, -a_0\}, \quad (3)$$
$$a_{III} = \max\{0, a_1\},$$
$$a_{IV} = \max\{0, -a_1\},$$

where $a_I, a_{II}, a_{III}, a_{IV} \in [0,1]$ represent the act of cable I, II, III, and IV shown in Figure 3-(E), respectively. The real robot module has only two chambers and shares the same actuation principle. The acts $a_I, a_{II} \in [0,1]$ of chambers in Figure 3-(F) are controlled by $a_0$ and linear to the chamber pressure ([0.08bar, 0.35bar]).

## 3.2. Simulation Experimental Results

We collect 16000 samples with the traditional random method and our method mentioned in Section 2.2 on a four-module simulation robot, as shown in Figure 4. It is evident that our method covers a larger space than the traditional method. The simulation experiment code can be found at https://github.com/zixichen007115/23ZCd.

Based on our dataset, we train three kinds of neural networks to estimate the actions and compare their accuracy. As shown in Figure 1-(C), the neural network input contains previous module states and actions, the module number label, and current states for training. The network output is the estimated actuation.

We compare our method with two methods applied in previous works[36], as shown in Figure S2. First, four LSTM networks are trained separately for four modules, and the ratios of the estimation errors to the actuation variable range are shown in Table 1. The estimation error is relatively high, which illustrates that it is challenging to estimate the actuation variables of one module without the knowledge of the other modules.

Then, we feed the input from all modules into one LSTM. Each LSTM unit takes input from all modules in one time step, and the units are connected according to the time sequence. The estimation errors are far lower than those of the 'four LSTM' method. Finally, we test our biLSTM and it achieves similar errors. Both LSTM and biLSTM take the information of all modules; hence, they achieve low errors.

In all the above approaches, the modules near the base have higher errors than the other modules, which means the base module is heavily influenced by the other modules. LSTM utilizes the recurrent structure to address the time sequence, while biLSTM addresses the space sequence. These two strategies can be combined to take both sequences into consideration, and this strategy may be included in our future work. Due to the low actuation estimation errors, we utilize LSTM and biLSTM to control the simulation modular robots with the same configuration trajectories.

In the simulation control task, we aim to control the configurations of four modules and indirectly achieve some end pose control targets. In task A, the modules are controlled to bend and rotate. In task B, the robot is desired to keep the end position invariant and change the orientation. In task C, the robot is desired to keep the end orientation invariant and change the position. The desired configuration trajectories are designed specifically after trial and error and can be found in the Supplementary Data. Accurate configuration design based on desired robot states like end pose or shape requires a planning strategy, which will be included in our future work.

The displacement between the real and desired module end unit vectors represents the configuration errors, and the ratios of module configuration errors to the unit vector length are shown in Table 2. The four-module robot motions in task A, B, and C with biLSTM are shown in Figure 5, and the desired and real module states are illustrated in Figure 6. The experimental videos are included in the Supplementary Movies. Both LSTM and biLSTM achieve low errors, and the maximal error of LSTM (7.62%) is higher than that of biLSTM(3.95%).

Then, we utilize biLSTM for six-module robot configuration control. Due to the modularity of this network, biLSTM trained with a four-module robot can also be leveraged on a six-module robot, but LSTM fails to transfer to a six-module robot because the LSTM unit input size is related to module number.

The configuration errors for the six-module simulation robot are shown in Table 3. The six-module robot motions with biLSTM are shown in Figure 7, and the desired and real module states are illustrated in Figure 8. The maximal error is 5.14%, demonstrating that biLSTM trained with a four-module robot can fulfill control tasks in a six-module robot. Considering the low errors of biLSTM on four-module and six-module simulation robots and its module number adaptability, we utilize biLSTM in real experiments.

### 3.3. Real Experimental Results

In real experiments, we first collect 15000 samples on a three-module robot following the data collection method proposed in Section 2.2 and train a biLSTM for control. The dataset is shown in Figure S5. The differences between two contiguous actuation variables in data collection are constrained to promise a smooth motion. The actuation variable estimation errors and variances of the first, second, and third modules are 3.12 ± 4.85%, 3.23 ± 5.23%, and 3.04 ± 4.38%.

Two tasks, 'edge' and 'down,' are designed. In 'edge,' the robot is controlled to reach the edge of the working space, and the actuation variables nearly reach the maximal actuation. In 'down,' the robot end is controlled downward during the motion, implying that the sum of the module bending angles is 0.

Each task has been carried out three times, and the configuration control errors for the three-module real robot are shown in Table 4. The module configuration is represented by the bending angle. The desired and real module states of the real three-module robot in task edge and down with biLSTM are shown in Figure 9. The experimental videos are included in the Supplementary Movies. The experimental results demonstrate that the biLSTM controller can control the real modular robot to follow trajectories with low errors (< 3°).

Then, considering that the module number in the simulation increases from four to six, we decrease the module number in the real experiments from three to two to prove the broad module number adaptability of the biLSTM configuration controller. The configuration control errors for the two-module real robot are shown in Table 4. The desired and real module states of the real two-module robot in task (A) edge and (B) down with biLSTM are shown in Figure 10.

The error of the second module in task 'edge' is 4.26°. Although it is not large considering the reachable space of this module (about 150°), it is the largest error in the real experiments. This may be caused by manufacturing errors among modules and can be solved by integrating online controllers[19]. Of note, in task 'down,' the bending angle of the first module in the two-module robot reaches about 30°, which is out of the reachable space of the first module in the three-module robot (< 6°). Even so, biLSTM trained with the three-module robot can fulfill this control task with relatively low errors, which shows that this controller is accurate and has module number adaptability.

## 4. Summary

This paper introduces a novel and accurate biLSTM configuration controller for module number adaptability and a data collection strategy specifically for modular soft robots. The dataset generated by the proposed collection strategy covers a larger space than that from the traditional method, and the biLSTM controller can decide actuation variables based on the previous and other module states. Simulation cable-driven robots with four and six modules and real pneumatic robots with three and two modules are leveraged for experiments, and the results have proven that the biLSTM controller for module number adaptability can fulfill module configuration control tasks on robots with different module numbers.

As far as we know, this is the first paper that includes biLSTM as a soft robot controller. The application of this network is inspired by the structure of modular soft robots shown in Figure 1-(A). To control a system with such an inner interaction relationship, we find a neural network sharing a similar interaction principle among units, which is biLSTM. This kind of strategy is a hint at applying data models that share similar principles with physical systems and may expand the neural network application in soft robots. For example, besides RNN, which can represent the time sequence of soft robot motion, a generative adversarial network (GAN) may be employed to train a pair of robot model and controller simultaneously, considering the pairing relationship of modeling and control.

As mentioned above, we may utilize biLSTM units to infer the space sequence and LSTM units to infer the time sequence in a large neural network in future work. Also, a planning approach may be introduced to map between the task and configuration space, as shown in Figure 1-(A). The planning method may combine neural networks and some physical or analytical models like [37]. Based on the proposed offline-trained controller, we can integrate it with an online controller like [19,38] to decrease the control errors and achieve interchangeability among modules.

## Acknowledgments

We acknowledge the support of the European Union by the Next Generation EU project ECS00000017 'Ecosistema dell'Innovazione' Tuscany Health Ecosystem (THE, PNRR, Spoke 4: Spoke 9: Robotics and Automation for Health.)

## Author Disclosure Statement

No competing financial interests exist.

# Tables

Table 1. Mean and standard derivation of actuation value estimation errors with different networks

|          | four LSTM        | LSTM           | biLSTM         |
|----------|------------------|----------------|----------------|
| module 1 | 18.52 ± 14.56%   | 0.93 ± 0.95%   | 1.63 ± 1.71%   |
| module 2 | 12.19 ± 10.33%   | 0.85 ± 0.85%   | 1.49 ± 1.52%   |
| module 3 | 7.69 ± 6.60%     | 0.76 ± 0.68%   | 1.11 ± 1.05%   |
| module 4 | 3.74 ± 3.21%     | 0.64 ± 0.57%   | 0.70 ± 0.65%   |

Table 2. Mean and standard derivation of configuration control errors for the four-module simulation robot

|          | LSTM-A        | LSTM-B        | LSTM-C        | biLSTM-A      | biLSTM-B      | biLSTM-C      |
|----------|---------------|---------------|---------------|---------------|---------------|---------------|
| module 1 | 4.70 ± 2.65%  | 2.07 ± 0.37%  | 7.62 ± 2.59%  | 3.95 ± 1.84%  | 0.72 ± 0.29%  | 3.61 ± 1.72%  |
| module 2 | 3.77 ± 1.73%  | 0.93 ± 0.49%  | 5.97 ± 1.69%  | 3.45 ± 1.09%  | 0.67 ± 0.28%  | 2.24 ± 1.36%  |
| module 3 | 1.29 ± 1.16%  | 1.58 ± 0.70%  | 2.21 ± 1.07%  | 1.64 ± 1.05%  | 1.16 ± 0.63%  | 1.01 ± 0.40%  |
| module 4 | 2.10 ± 1.41%  | 1.89 ± 0.91%  | 1.85 ± 1.00%  | 1.09 ± 0.81%  | 1.92 ± 0.78%  | 1.18 ± 0.59%  |

Table 3. Mean and standard derivation of configuration control errors for the six-module simulation robot

|          | A             | B             | C             |
|----------|---------------|---------------|---------------|
| module 1 | 2.94 ± 1.51%  | 3.99 ± 0.43%  | 1.82 ± 0.25%  |
| module 2 | 1.73 ± 0.94%  | 2.43 ± 0.42%  | 2.28 ± 0.40%  |
| module 3 | 5.18 ± 1.74%  | 0.65 ± 0.33%  | 3.35 ± 0.66%  |
| module 4 | 5.14 ± 1.84%  | 1.80 ± 0.67%  | 1.21 ± 0.63%  |
| module 5 | 3.37 ± 0.97%  | 4.30 ± 0.57%  | 3.30 ± 0.75%  |
| module 6 | 2.49 ± 0.45%  | 1.80 ± 0.85%  | 3.58 ± 1.11%  |

Table 4. Mean and standard derivation of configuration control errors for the three-module and two-module real robot

|          | 3-edge         | 3-down         | 2-edge         | 2-down         |
|----------|----------------|----------------|----------------|----------------|
| module 1 | 0.58 ± 0.45°   | 0.50 ± 0.47°   | 0.94 ± 0.63°   | 1.47 ± 0.86°   |
| module 2 | 0.92 ± 0.76°   | 1.72 ± 1.19°   | 4.26 ± 2.80°   | 1.69 ± 1.28°   |
| module 3 | 2.69 ± 1.66°   | 1.98 ± 1.52°   | /              | /              |

# Figures

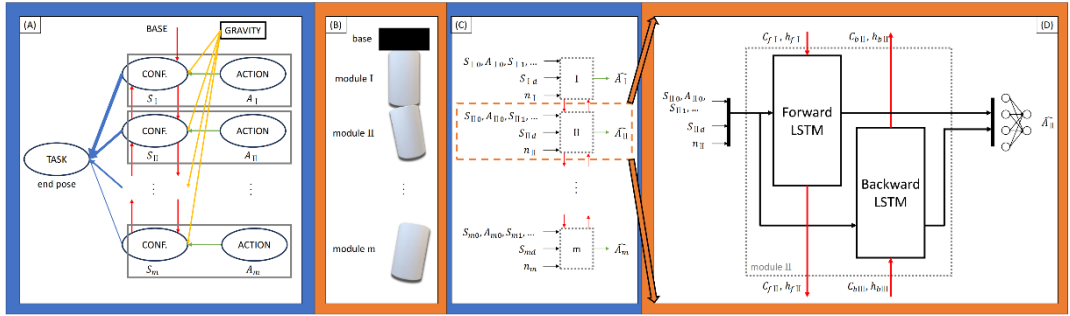

Figure 1. (A) Modular soft robot structure. The actuation of each module $A$ in the action space will affect its configuration $S$ in the configuration space. Also, the configuration is affected by gravity, interaction with the adjacent modules, and base. Finally, all the configurations have different impacts on the end pose in the task space. This paper proposes a configuration controller, shown as the boxes. (B) The diagram of a modular soft robot. (C) BiLSTM controller. Each unit takes the desired module states $S_d$, module number label $m$, previous module states and actuations, $S$ and $A$, as input and produces the actuation $\hat{A}$ for this time step. (D) The diagram of the biLSTM unit for module II.

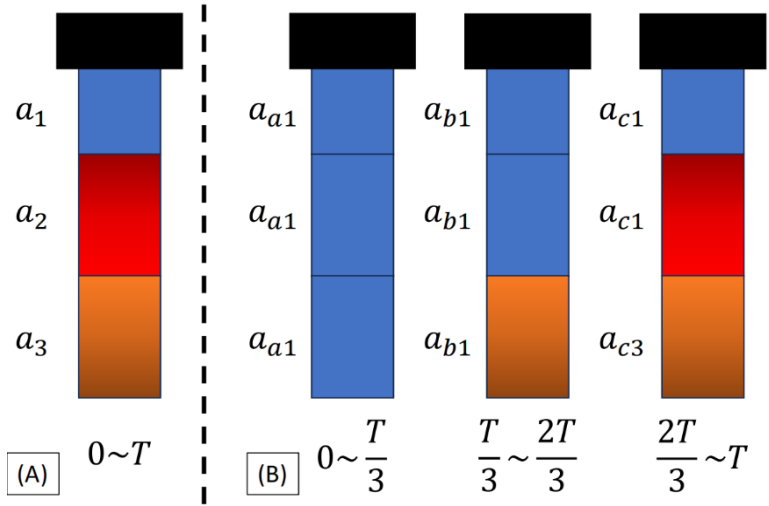

Figure 2. (A) General data collection method. (B) Data collection method proposed for modular robots. $a_*$ represents different random actuation sequences.

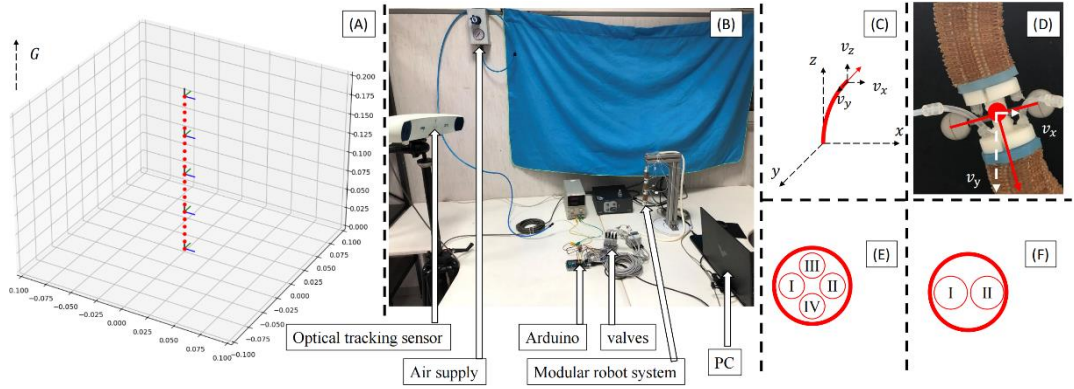

Figure 3. (A) Simulation robot diagram. The robot comprises four modules, and the end of each module is shown by three coordinate axes. The direction of gravity is upward in this diagram. (B) Real robot setup. A pneumatic robot composed of three modules is actuated by valves controlled by Arduino and PC. The optical tracking system tracks the optical tracker motion and sends to PC for biLSTM control. (C) Simulation and (D) real module configuration. The configuration of each module is shown by the orientation of end unit vector. (E) Simulation and (F) real module actuation. Each simulation module is actuated by four cables while the real one is actuated by two chambers.

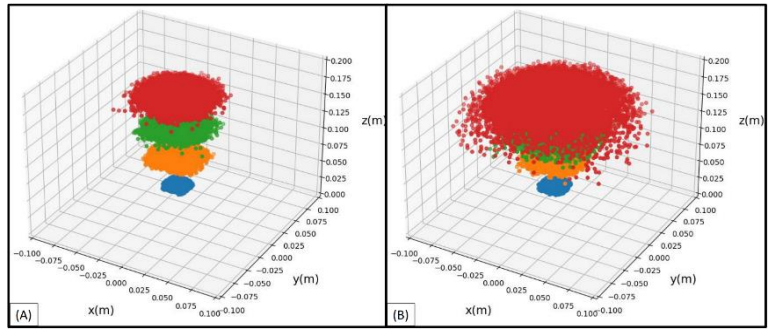

Figure 4. The diagram of datasets collected with (A) the traditional method and (B) our method. The points represent the end positions of the first, second, third, and final modules.

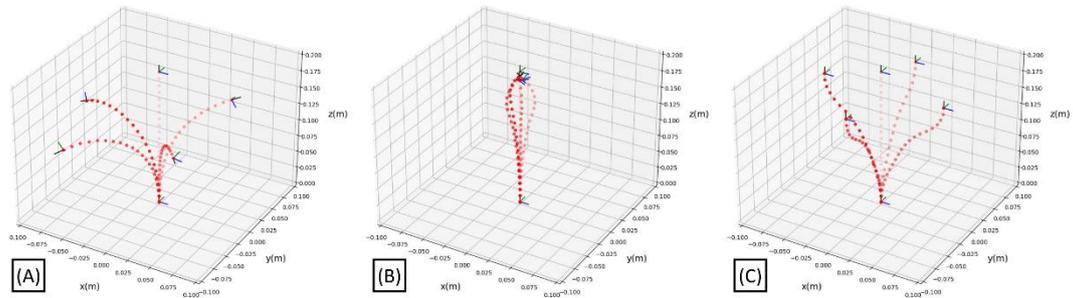

Figure 5. Simulation four-module robot motion in task A, B, and C with biLSTM. The color becomes deep with the improvement of time steps.

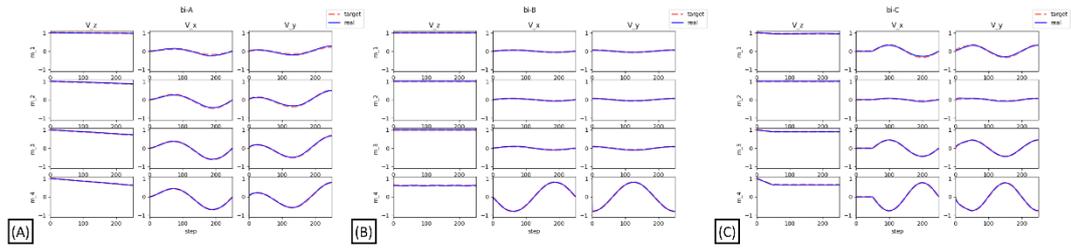

Figure 6. The desired and real module states of the simulation four-module robot in task A, B, and C with biLSTM.

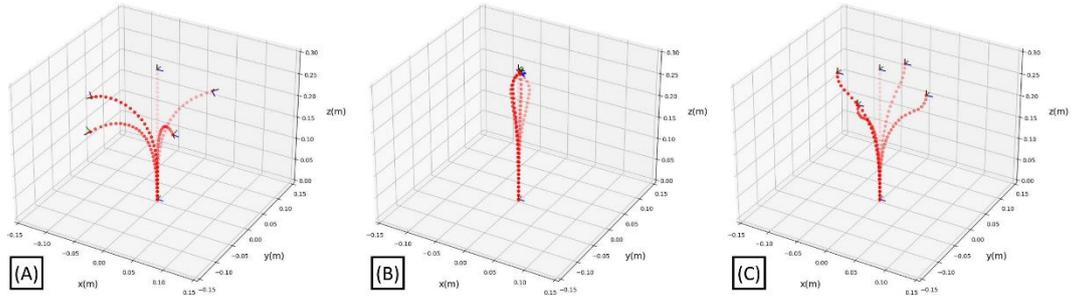

Figure 7. Simulation six-module robot motion in task A, B, and C with biLSTM. The color becomes deep with the improvement of time steps.

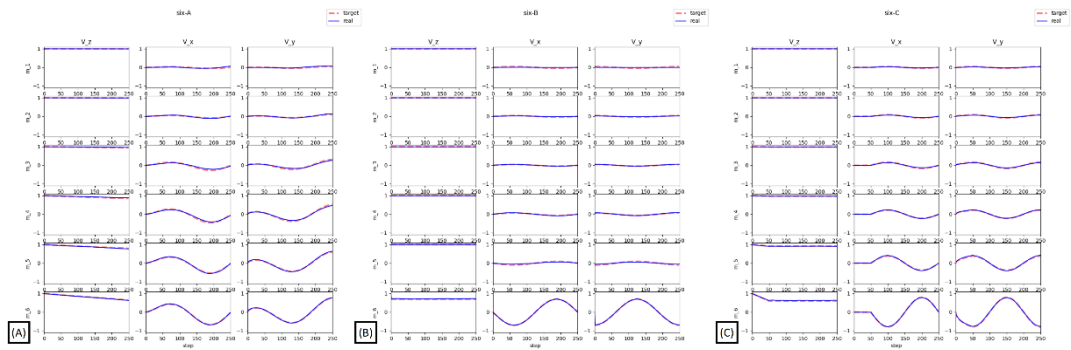

Figure 8. The desired and real module states of the simulation six-module robot in task A, B, and C with biLSTM.

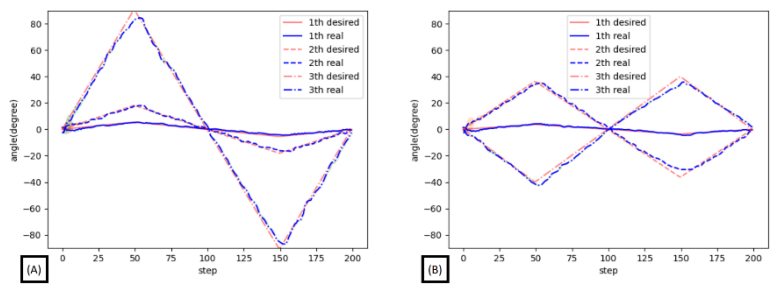

Figure 9. The desired and real module states of the real three-module robot in task (A) edge and (B) down with biLSTM.

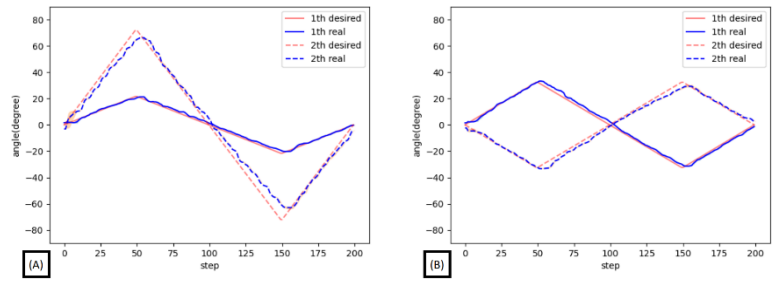
Figure 10. The desired and real module states of the real two-module robot in task (A) edge and (B) down with biLSTM.

# Supplementary Data

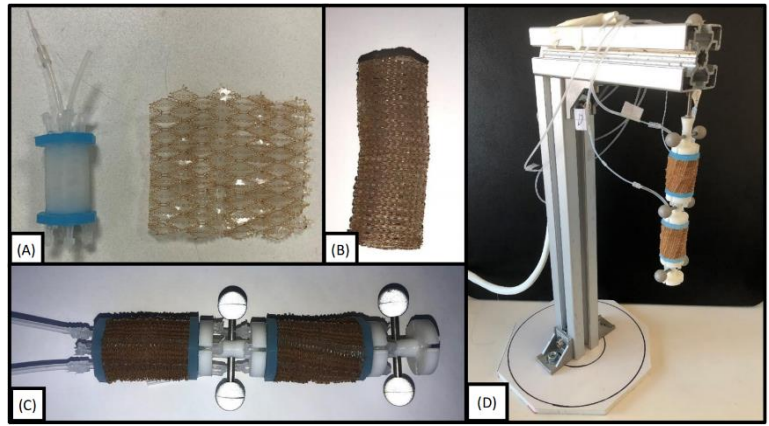

Figure S1. (A) Pneumatic robot made of silicone and apart of unfolded origami structure. (B) a folded origami structure. (C) Each origami structure covers one pneumatic robot. One 3D-printed connector connects two modules, and a pair of optical trackers are fixed on the connector. (D) A modular robot system is connected to a metal stick, and three fixed optical trackers are applied for system calibration.

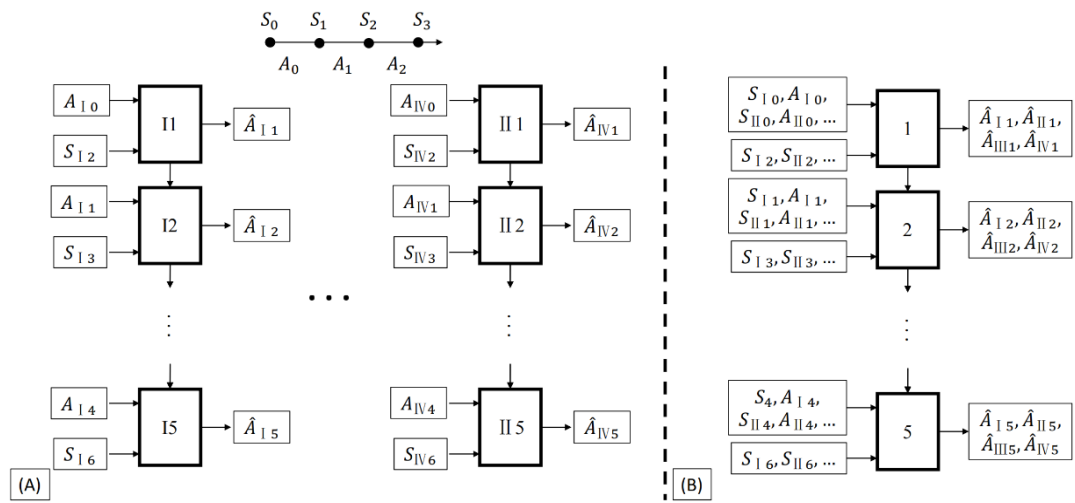

Figure S2. The diagram of (A) four LSTM and (B) LSTM.

Table S1. Neural networks parameters

|  | Four LSTM | LSTM | biLSTM | biLSTM (real) |
|---|---|---|---|---|
| layer number | 4 | 4 | 4 | 4 |
| hidden state size | 128 | 128 | 128 | 128 |
| time step number | 5 | 5 | 5 | 10 |
| batch size | 64 | 64 | 64 | 128 |
| optimizer | Adam | Adam | Adam | Adam |
| Learning rate | 0.001 | 0.001 | 0.001 | 0.0003 |

**Simulation configuration trajectory:**

In task A, the trajectories for $v_x$, $v_y$, and $v_z$ are

$$\begin{cases} v_{dz}(t) = 1 - (1 - v_{zmin}) * \dfrac{t}{t_{max}} \\ v_{dx}(t) = \sin\left(2\pi * \dfrac{t}{t_{max}}\right) * \sqrt{1 - v_{dz}(t)^2} \\ v_{dy}(t) = \cos\left(2\pi * \dfrac{t}{t_{max}}\right) * \sqrt{1 - v_{dz}(t)^2} \end{cases} \quad (S1)$$

where $v_{zmin}$ denotes the minimal value of $v_z$. It is [0.975,0.850,0.725,0.625] for four-module robot and [0.998, 0.995, 0.950, 0.850, 0.800, 0.650] for six-module robot. $t_{max}$ represents the length of each control trial and is 250 in simulation.

In task B, the trajectories for $v_x$, $v_y$, and $v_z$ are

$$\begin{cases} v_{dz}(t) = v_{dz} \\ v_{dx}(t) = a * \sin\left(2\pi * \dfrac{t}{t_{max}}\right) * \sqrt{1 - v_{dz}(t)^2} \\ v_{dy}(t) = a * \cos\left(2\pi * \dfrac{t}{t_{max}}\right) * \sqrt{1 - v_{dz}(t)^2} \end{cases} \quad (S2)$$

where $v_{dz}$ is [0.998,0.998,0.996,0.600] for four-module robot and [0.999, 0.999, 0.999, 0.998, 0.995, 0.708] for six-module robot. $a$ is employed to set the module rotation direction. It is [1,1,1,-1] for four-module robot and [1,1,1,1,-1,-1] for six-module robot.

In task C, the trajectories for $v_x$, $v_y$, and $v_z$ are

$$\begin{cases} v_{dz}(t) = 1 - (1 - v_{zmin}) * \dfrac{t}{50} \\ v_{dx}(t) = 0 \\ v_{dy}(t) = a * \sqrt{1 - v_{dz}(t)^2} \end{cases}, t < 50, \quad (S3)$$

$$\begin{cases} v_{dz}(t) = v_{zmin} \\ v_{dx}(t) = a * \sin\left(2\pi * \dfrac{t-50}{200}\right) * \sqrt{1 - v_{dz}(t)^2} \\ v_{dy}(t) = a * \cos\left(2\pi * \dfrac{t-50}{200}\right) * \sqrt{1 - v_{dz}(t)^2} \end{cases}, t \geq 50, \quad (S4)$$

where $v_{zmin}$ is [0.941,0.998,0.897,0.650] for four-module robot and [0.999, 0.996, 0.985, 0.975, 0.925, 0.600] for six-module robot. $a$ is [1,1,1,-1] for four-module robot and [1,1,1,1,1,-1] for six-module robot.

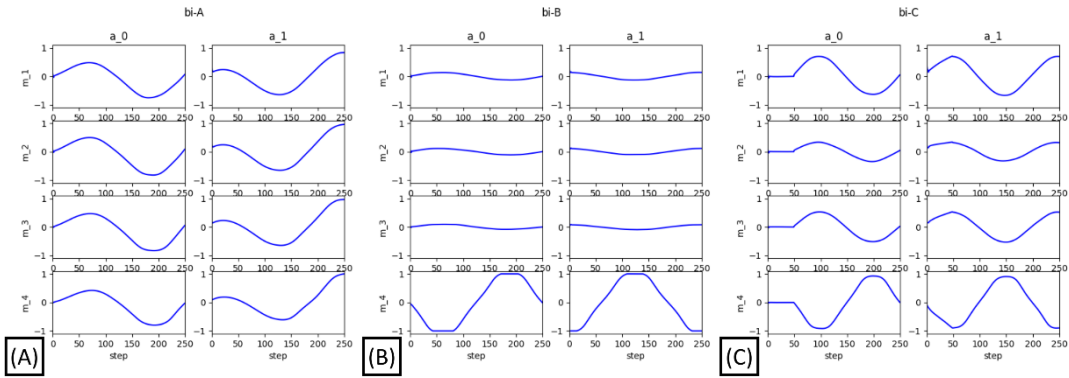

Figure S3. Actuation variables of simulation four-module robot in task A, B, and C with biLSTM.

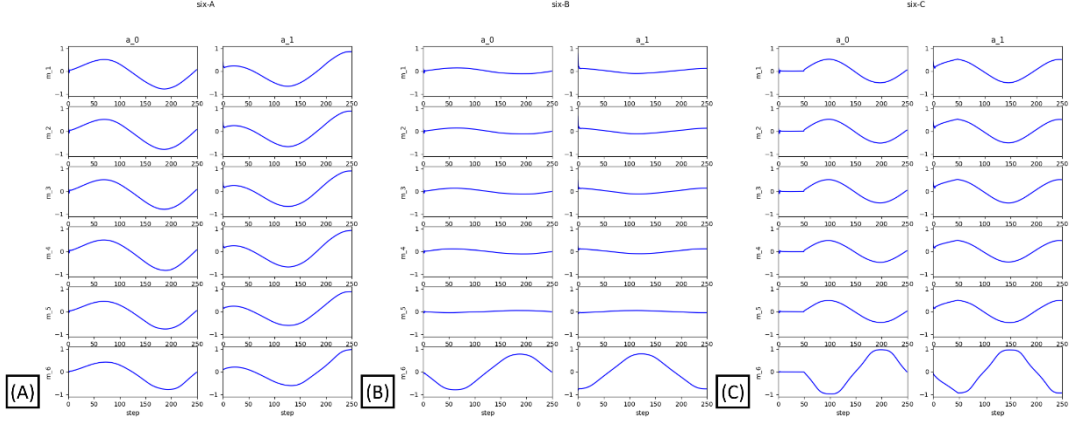

Figure S4. Actuation variables of simulation six-module robot in task A, B, and C with biLSTM.

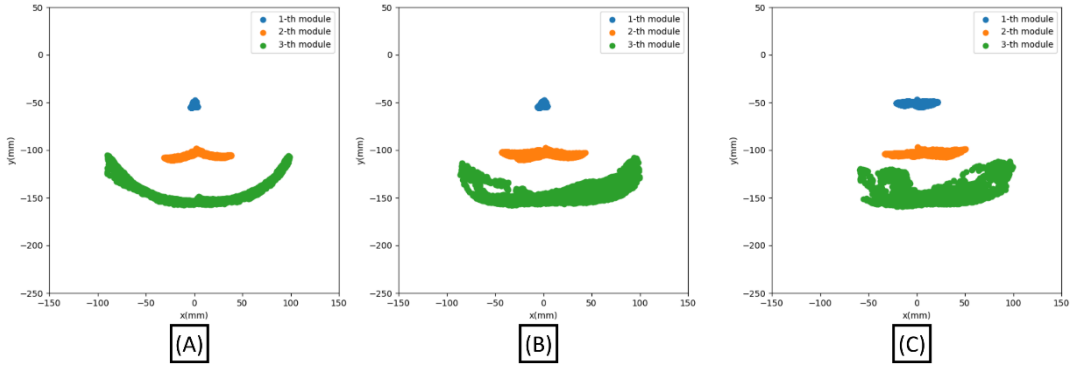

Figure S5. Dataset collected with our proposed data collection method. (A), (B) and (C) represent three actuation methods shown in Figure 2-(B).

**Real angle trajectory:**

In real experiments, each trial contains 200 steps, which means $t_{max} = 200$.

The desired angle trajectories are

$$ang_d = \begin{cases} ang_{max} * \dfrac{t}{50}, & t < 50 \\ ang_{max} * \left(2 - \dfrac{t}{50}\right), & 50 \leq t < 150, \\ ang_{max} * \left(\dfrac{t}{50} - 4\right), & 150 \leq t \end{cases} \quad (S5)$$

where $ang_{max}$ is the maximal module bending angle. For task edge, it is [5.4°, 18°, 90°] for three-module robot and [21.6°, 72°] for two-module robot. For task down, it is [3.6°, 36°, -39.6°] for three-module robot and [3.24°, -32.4°] for two-module robot.

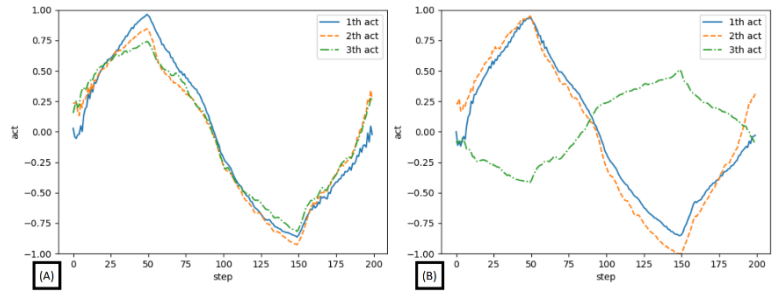

Figure S6. Actuation variables of real three-module robot in task (A) edge and (B) down with biLSTM.

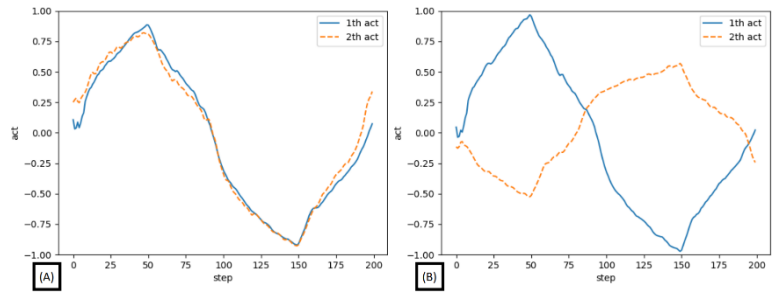

Figure S7. Actuation variables of real two-module robot in task (A) edge and (B) down with biLSTM.